\pdfoutput=1

\documentclass[11pt]{article}
\usepackage[dvipsnames]{xcolor}

\usepackage[preprint]{acl}

\usepackage{times}
\usepackage{latexsym}
\usepackage{xurl}

\usepackage[T1]{fontenc}

\usepackage[utf8]{inputenc}

\usepackage{microtype}

\usepackage{inconsolata}

\usepackage{graphicx}

\usepackage{booktabs}
\usepackage{amsmath}
\usepackage{multirow, bigdelim}
\usepackage{makecell}
\usepackage{graphicx}
\usepackage{adjustbox}

\usepackage{listings}
\usepackage{color}

\definecolor{mygreen}{rgb}{0,0.6,0}
\definecolor{mygray}{rgb}{0.5,0.5,0.5}
\definecolor{mymauve}{rgb}{0.58,0,0.82}

\newcommand{\tightparagraph}[1]{\paragraph{#1}}

\usepackage{mdframed}

\usepackage{rotating}

\makeatletter
\newcommand\footnoteref[1]{\protected@xdef\@thefnmark{\ref{#1}}\@footnotemark}
\makeatother

\usepackage{array}

\usepackage{enumitem}
\newlist{steps}{enumerate}{1}
\setlist[steps, 1]{label = Step \arabic*:, left=2em}

\usepackage{tikz}
\usetikzlibrary{shapes.geometric, arrows}

\tikzstyle{generated_summary} = [rectangle, minimum width=2cm, text width=2cm, minimum height=0.75cm,text centered, draw=black, fill=blue!30]
\tikzstyle{reference_summary} = [rectangle, minimum width=2cm, text width=2cm, minimum height=0.75cm,text centered, draw=black, fill=orange!30]
\tikzstyle{reference_code} = [rectangle, minimum width=2cm, text width=2cm, minimum height=0.75cm,text centered, draw=black, fill=orange!50]

\tikzstyle{LLM} = [rectangle, minimum width=3cm, minimum height=0.6cm,text centered, draw=black, fill=gray!20, line width=0.5mm]
\tikzstyle{process} = [rectangle, minimum width=2cm, minimum height=0.4cm,text centered, draw=black, fill=white]

\tikzstyle{arrow} = [thick,->,>=stealth]
\tikzstyle{dotted_arrow} = [dashed, ->, >=stealth]

\usepackage{pgfplots}
\pgfplotsset{compat=1.18}

\usepackage{pifont}
\newcommand{\cmark}{\ding{51}}%
\newcommand{\xmark}{\ding{55}}%

\usepackage{subfig}

\usepackage{float}

\title{Simple and Effective Baselines for Code Summarisation Evaluation}

\author{Jade Robinson \and Jonathan K. Kummerfeld \\
  University of Sydney \\
  \texttt{jonathan.kummerfeld@sydney.edu.au}}

\begin{document}
\maketitle
\begin{abstract}
Code documentation is useful, but writing it is time-consuming.
Different techniques for generating code summaries have emerged, but comparing them is difficult because human evaluation is expensive and automatic metrics are unreliable.
In this paper, we introduce a simple new baseline in which we ask an LLM to give an overall score to a summary.
Unlike n-gram and embedding-based baselines, our approach is able to consider the code when giving a score.
This allows us to also make a variant that does not consider the reference summary at all, which could be used for other tasks, e.g., to evaluate the quality of documentation in code bases.
We find that our method is as good or better than prior metrics, though we recommend using it in conjunction with embedding-based methods to avoid the risk of LLM-specific bias.
\end{abstract}

\section{Introduction}

Relevant and up-to-date documentation is useful for software maintenance \cite{stapleton2020, misra2020, desouza2006}/
To support one important form of documentation, researchers have developed models that generate one-line summaries of functions \citep[inter alia]{hudeepcom2018, leclair2020, nguyen2024}.
However, evaluating these models is difficult.
Expert human evaluations are expensive, slow to collect, and hard to consistently reproduce.
Automatic metrics address are cheap and consistent, but they have weak-to-moderate correlation with human scores \cite{roy2021, haque2022, mastropaolo2024}.

In this paper, we introduce a simple baseline: directly querying an LLM to get an overall rating of a generated summary.
This approach considers the code when judging the summary, which most current metrics do not.
We also propose a reference-free variant, which has not previously been done for this task.
Not needing a reference summary enables new uses of these metrics, such as to flag low quality summaries in a code base or as part of the summary generation process.

We compare with all of the standard n-gram based metrics, a model-based metric \cite{mastropaolo2024}, and embedding-based metrics.
We evaluate by measuring correlation with two datasets of human judgements \cite{haque2022, roy2021}.
In appendices, we also provide results on two datasets that consider specific aspects of summary quality.

Our approach is the best at predicting an overall score.
For similarity with a reference, there is no significant difference between our approach and alternatives.
We do find a risk that our method prefers output if it comes from the same LLM as the metric, and so we recommend using our method alongside an embedding-based metric.

While evaluation by querying an LLM has been done in other tasks with natural language outputs, our results differ from work in other areas.
For example, unlike in machine translation, our method remains just as effective without a reference, and it improves over a metric using a supervised model, and unlike in QA, our method does not favour longer (or shorter) summaries.
These differences highlight the distinctiveness of code summarisation and therefore, the value of research in this space.
Our work provides novel baselines that are simple and effective, forming a solid foundation for further exploration.

\section{Related Work}

\paragraph{Code Summarisation Evaluation}
N-gram metrics, such as BLEU, METEOR, and ROUGE-L, were the first approach for evaluation, but have low correlation with human evaluation \citet{roy2021}.
Embedding-based approaches, such as SentenceBERT, improve on n-gram metrics, but still have a weak-to-moderate correlation \cite{haque2022, mastropaolo2024}. 
One trained metric exists, SIDE, and improves slightly over embedding methods \cite{mastropaolo2024}.

Despite these findings, research still relies on n-gram metrics for evaluation. Of ten new code summarisation papers in 2024 \cite{nguyen-etal-2024-hierarchynet, su2024distilgpt, su2024contextawarecodesummarygeneration, zhao2024, li2024, pan2024, sun2024, ahmed2024, cai2024, mao2024}, six used only n-gram metrics, three used n-gram metrics and embedding-based metrics, and one only used human evaluation.

\paragraph{Human Evaluation Datasets}
We focus on two datasets that were collected specifically for code summarisation metric evaluation \cite{roy2021, haque2022}.
We also draw data from papers that proposed new code summarisation methods and asked people to evaluate specific aspects of quality
\cite{gao2023, su2024contextawarecodesummarygeneration}.
Those results are mentioned in analysis and included in appendices due to space constraints.

\paragraph{LLM-prompting based NLG Evaluation}
Prompting has been successfully used to evaluate other forms of Natural Language Generation, e.g., for text summarisation and dialogue generation \cite{liu-etal-2023-g}, and machine translation \cite{kocmi-federmann-2023-large}.
We observe some key differences between our results and other NLG work.
We achieve equally strong results without a reference, but \citet{qian-etal-2024-large} and \citet{huang-etal-2024-lost} investigate different prompting techniques and find that the reference summary is very beneficial.
We also find that our approach consistently improves over a trained method, while trained models are still the most effective for MT \cite{anugraha-etal-2024-metametrics,freitag-etal-2024-llms}, probably because of the larger and higher quality datasets for metric development in MT.

There has also been considerable work evaluating the potential biases of LLM evaluators \cite{wu2023stylesubstanceevaluationbiases, zheng2024, koo-etal-2024-benchmarking}, finding evidence that LLMs tend to evaluate their own outputs more highly and favour longer responses. We investigate this issue in Section~\ref{analysis}. 

\paragraph{Reference-Free Metrics}
We introduce the first reference-free approach for code summarisation evaluation, but there is significant prior work for other tasks \cite{rei-etal-2021-references,scialom-etal-2021-questeval}.
These often have better correlations with human evaluations than equivalent reference-based metrics. 
However, \citet{deutsch-etal-2022-limitations}  argue that reference-free metrics are essentially creating their own pseudo-references, and so are constrained by their own generation ability.
We agree that reference-free metrics are not a complete substitute, but for code summarisation they have the additional benefit that they could be used to flag low quality summaries within an existing code base.

\section{Task}

Code summarisation is the task of generating a summary of a code snippet.
We are proposing new metrics for this task.
The aim of the metric is to output a score that captures the overall quality of the summary, so that it can provide a broad indicator of the model's performance.
These metrics have access to the code, the generated summary, and a human-written reference summary.
However, we will also consider a variant of our approach that does not use the reference.
We measure the quality of the metric by looking at how well it correlates with human ratings of overall score and similarity.

\section{New Metric: Ask LLM Directly}

Our metric is simple: ask an LLM to give the summary a rating, just like asking a human.
One benefit is that this approach can consider the relevant \textit{code} as well as the reference summary.
In contrast, n-gram and embedding based metrics only measure the similarity between the generated summary and a reference summary.
Our metric can also work without a reference.
We include this variant in our results and note that (1) it is useful when high-quality references are not available, and (2) it could be used outside of model evaluation, for example to identify low quality human-written documentation.

To develop this metric we tested different techniques such as chain-of-thought reasoning, role-based prompting and varying the problem description.
We also considered question-answering based prompts, where we focused on whether the LLM was able to answer questions about the reference using information from the generated summary.
For details, see \autoref{appendix:variations}.

\section{Experiments}

\subsection{Datasets}

We use two datasets that were created for metric evaluation.
We aim to produce a single score, and so the most relevant data is \citet{roy2021}'s Overall Score, a direct assessment of the overall quality of the summary.
We also consider \citet{haque2022}'s Similarity, which measures the similarity with the reference, but that does not account for a high quality but different summary.
To avoid overfitting, during development we used a subset of the data.
For the final results we used all of the data with 10-fold cross-validation. 

In analysis, we also consider human evaluations of Adequacy that were collected in the process of evaluating a code summarisation system \cite{gao2023}.
Additional details are in Appendix~\ref{sec:data-measures} and results comparing with specific aspects of quality are in Appendix~\ref{sec:other-aspects}.

We release a version of all the datasets reformatted to be consistent, and with all of the same information.
This was somewhat involved as some datasets did not directly include the code.
Fortunately, they did indicate their code and documentation source, and so we could go back to that source and match the summary to find the code.

\subsection{Measuring Correlation}

As in previous papers which evaluate code summarisation metrics \cite{roy2021, haque2022, mastropaolo2024}, we aim to maximise correlation with human evaluation scores.
We follow \citet{haque2022}'s methodology:
(1) when there are multiple human scores for a sample, we compare with the mean to reduce the impact of noise from disagreement, and
(2) we use Spearman's Rank correlation for each metric because, unlike Pearson's correlation, it does not assume a normal distribution.
We use a permutation test for significance testing, see Appendix~\ref{sec:significance} for details.

\subsection{Metrics}

We consider the most commonly used metrics (BLEU, METEOR and ROUGE-L), the best metrics according to prior work (SIDE and SentenceBERT), two new embeddings (gite-base-en, and coyage-code-3), and our own metric (ask-LLM and ask-LLM-no-ref), where \emph{LLM} is the name of the model that is queried, and \emph{no-ref} indicates the variant in which no reference summary is provided in the prompt.
For further details, see Appendix~\ref{sec:reproduction instructions}.
Metrics that are evaluated here for the first time are in italics in Table~\ref{overall-score-table}.

\section{Results}

\begin{table}
    \centering
    \small
    \renewcommand{\arraystretch}{0.9}
    \setlength{\tabcolsep}{1pt}
    \begin{tabular}{@{}r@{\,}lcc@{}}
    \cmidrule[\heavyrulewidth]{2-4}
    & & Overall Score & Similarity \\
     & Metrics &(\citeauthor{roy2021}) & (\citeauthor{haque2022})\\
    \cmidrule{2-4}
    \ldelim\{{3}{*}[n-gram]
    & BLEU-A & 0.28 & 0.55 \\
    & METEOR & 0.31 & 0.75\\
    & ROUGE-L & 0.21 & 0.47  \\
    \cmidrule{2-4}
    \ldelim\{{1}{*}[trained] & SIDE & 0.38 & 0.32 \\
    \cmidrule{2-4}
    \ldelim\{{3}{*}[embedding] & SentenceBERT & 0.36 & 0.76\\
     & \textit{gte-base-en} & 0.38 & 0.80\\
     & \textit{voyage-code-3} & 0.43 & \textbf{0.81}\\
    \cmidrule{2-4}
    \ldelim\{{5}{*}[ask-LLM] & \textit{ask-OLMo} & 0.35 & 0.49 \\
     & \textit{ask-OLMo-no-ref} & 0.36 & 0.61 \\
     & \textit{ask-gpt} & 0.42 & 0.48\\
     & \textit{ask-claude} & \textbf{0.47} & 0.57\\
     & \textit{ask-claude-no-ref} & 0.46 & 0.61\\
    \cmidrule[\heavyrulewidth]{2-4}
    \end{tabular}
    \caption{Spearman's Correlation with Human Ratings for Overall Score and Similarity}
    \label{overall-score-table}
\end{table}

Table~\ref{overall-score-table} shows correlations with Overall Score and Similarity to the reference summary.
Below, we note several key results.

\tightparagraph{N-gram metrics are not as effective.}
For Overall Score, the trained method (SIDE), the best embedding-based approach (voyage-code-3) and the best ask-LLM approach (ask-claude) outperform the best n-gram metric (BLEU-A).
All of these improvements are statistically significant according to a permutation test at below the $0.01$ level\footnote{Specific p-values are included in \autoref{sec:significance}}.
For Similarity, we find a different pattern, with SIDE performing worst, and the other three types of metrics in similar ranges.
We find no statistically significant difference between the best n-gram based metric (METEOR) and either the best embedding-based metric (voyage-code-3) or the best ask-LLM metric (ask-claude-no-ref).

\tightparagraph{Embedding metrics are comparable to ask-LLM metrics.}
On Overall Score, the best embedding-based approach (voyage-code-3) and the best ask-LLM approach (ask-claude) are not statistically significantly different.
For Similarity they are, with the embeddings being better, but we would expect embeddings to be better suited to that task.
Note in particular that a summary may be good even though it isn't similar to the reference, and so a metric that focuses on similarity will sometimes struggle.
There is also the issue that Similarity is only a measure of quality if the reference is high quality.
In code summarisation datasets, nearly all reference summaries are stripped from Github with limited manual oversight.
This introduces many quality issues.

\tightparagraph{Newer embeddings are better.}
For both Overall Score and Similarity, the newest embedding based metric, using voyage-code-3, improves on the previous state-of-the-art embeddings-based metric SentenceBERT.
This is good news, since it indicates that continued progress on embedding methods is likely to continue to provide improvements here.
One key difference between these approaches is cost, which will be discussed below.

\tightparagraph{ask-LLM-no-ref is just as effective.}
The performance of the Ask-LLM-Directly style metrics is stable regardless of whether the reference summary is provided, with no statistically significant difference between the two.

\tightparagraph{Different LLMs may perform differently.}
The choice of model (e.g. OLMo vs Claude) does lead to a significant difference.
However, we used Claude when fine-tuning our prompt, making it an unfair comparison.

\subsection{Analysis}
\label{analysis}

To understand the strengths and weaknesses of our approach, we conducted several additional experiments.

\paragraph{Ask-LLM method can't easily be adapted to different quality dimensions} 
\autoref{fig:quality-dim-variation} shows the results of our attempts to get the LLM to focus on specific aspects of quality.
We see very little variation, with the scores continuing to mainly reflect Adequacy.
Looking at specific examples, we found two issues.
First, mentioning unrelated issues, e.g., for conciseness it produced: ``The generated summary contains \textit{incorrect information} and does \textit{not accurately} summarize the function.''.
Second, inconsistency, e.g., for conciseness it produced  ``...the \textit{lack of specificity} makes the generated summary less informative ...''. 
We did not explore this further since our focus is on a single metric that aligns with overall quality.
However, note that full results in the appendices show that despite these issues, correlation with specific aspects was better than prior methods.

\begin{table}
\centering
\small
\renewcommand{\arraystretch}{0.9}
\begin{tabular}{@{} p{3.4cm} c c c @{}}
\toprule
 \multicolumn{1}{c}{} & \multicolumn{3}{c}{\citeauthor{gao2023} - Java (Train)} \\
\cmidrule(lr){2-4} 
 \multicolumn{1}{c}{Quality Dimensions} & Adequate & Concise & Fluent \\ 
\midrule
 \textbf{Accuracy:} ``Independent of other factors, I feel the new summary is accurate''& 0.59 & \textbf{0.38} & \textbf{0.44} \\ 
\midrule
 \textbf{Adequacy:} ``The new summary contains all of the important information required for understanding the method''& \textbf{0.60} & 0.35 & 0.37 \\ 
\midrule
  \textbf{Conciseness:} ``The new summary only contains necessary information.''& 0.59 & 0.35 & 0.41 \\ 
\bottomrule
 \end{tabular}
\caption{Test of Different Quality Dimensions}
\label{fig:quality-dim-variation}
\end{table}

\paragraph{Ask-LLM is Not Sensitive to Length}
Many studies suggest that LLM evaluators are biased towards longer outputs \cite{wu2023stylesubstanceevaluationbiases, zheng2024, koo-etal-2024-benchmarking}.
However, for our metric, looking at the scores assigned by different metrics and the number of characters in the generated summary, in most cases we find the correlation is close to zero.
For full results, see Appendix~\ref{sec:corr-comm-leng}.

\paragraph{Model Sensitivity}
There is a risk that an LLM will prefer its own output.
We considered the relative ranking of each model according to each metric.
Surprisingly, ask-gpt rates the models that use GPT-4 as the worst overall.
None of the data we had used Claude and so we generated our own summaries with Claude and valuated them.
While Claude did find some issues within summaries it had generated, in 92.7\% of cases it gives its summaries the highest possible rating.
For full results, see Appendices~\ref{sec:claude-evaluates-itself} and \ref{sec:rel-ranks}.
Based on this, we recommend using these metrics in combination with embedding based methods.

\paragraph{Costs}
Table~\ref{tab:costs} shows the cost per summary of each of the metrics.
These are API costs for commercial tools and compute costs open source model OLMo-2 (we used an A100).
These results show that these approaches are clearly much cheaper than running human evaluations, but still more expensive than metrics which can be run locally, e.g. gte-base-en, Sentence-BERT and n-gram methods.

\begin{table}
    \centering
    \small
    \renewcommand{\arraystretch}{0.9}
    \begin{tabular}{lc}
         \toprule
         Metrics & Cost per Query (USD) \\
         \midrule
         voyage-code-3 & 0.000002 \\
         ask-OLMo/ask-OLMo-no-ref & 0.011 \\
         ask-gpt & 0.012\\
         ask-claude/ask-claude-no-ref & 0.024\\
         \bottomrule
    \end{tabular}
    \caption{Metric Costs}
    \label{tab:costs}
\end{table}

\section{Conclusion}

We introduce a simple LLM-based evaluation metric and evaluate it on the standard code summarisation datasets.
Our approach is consistently better than prior metrics.
We describe a reference-free variant of our approach, which also performs well, and could be used in a variety of ways.
We recommend that future work uses a combination of embedding and Ask-LLM metrics for development, and turn to human raters for final evaluation.
That will enabler faster development while maintaining reliable evaluation.

\section*{Limitations}

\paragraph{Languages Evaluated Against} Due to the lack of available data, it was not possible to get any human evaluation data from programming languages apart from Python and Java.
Of those, overall scores are only available for the Java-based datasets, though in our appendices we do include results for other scores on the Python datasets. 
Metrics that work for one language may not be as effective for another.
This means that our results may not apply as well across different languages that we weren't able to evaluate against. 
In particular, our method may be less effective on languages that are less widely used and so less well understood by LLMs.
However, by definition, rare langauges are not as common as common ones, and so our method will be useful for the vast majority of developers.

\paragraph{Reliance on Intrinsic Human Assessment} All of the studies we use ask raters to assess the quality of the summary directly, rather than assess the impact of the different summaries on downstream tasks. As such, the ratings may not be optimising for summaries that aid development but rather developer perception, which while likely a good proxy, will never be perfect.
The reference summaries are also human-written and vary considerably in quality.
For reference-based methods, that could be misleading, as being similar to the reference may not indicate a summary is good.

\paragraph{Prompting Approaches Tested} We were not able to run every prompt variation for every model, GPT and OLMo generally seem to perform worse here but this could be because the prompts were decided using Claude, but due to budget restraints we only tested the best performing prompt from initial testing on Claude. 

\section*{AI Assistance Statement}
ChatGPT was used to ask LaTeX questions to assist with the formatting of this paper.  

\section*{Acknowledgments}

This material is partially funded by an unrestricted gift from Google, and by the Australian Research Council through a Discovery Early Career Researcher Award.

\bibliography{custom}

\newpage

\appendix

\section{Other Aspects of Quality}
\label{sec:other-aspects}

\begin{table*}[t]
    \centering
    \small
    \renewcommand{\arraystretch}{0.9}
    \setlength{\tabcolsep}{1pt}
    \begin{tabular}{@{}r@{\,}lccccc@{}}
    \cmidrule[\heavyrulewidth]{2-7}
    & & Content & & & Adequacy & Adequacy \\
    & & Adequacy & Adequacy & Accuracy & (Java) & (Python) \\
     & Metrics &(\citeauthor{roy2021}) & (\citeauthor{haque2022}) & (\citeauthor{haque2022}) &(\citeauthor{gao2023}) & (\citeauthor{gao2023}) \\
    \cmidrule{2-7}
    \ldelim\{{3}{*}[n-gram]
    & BLEU-A & 0.27 & 0.37 & 0.33 & 0.48 & 0.27 \\
    & METEOR & 0.31 & 0.45 & 0.47 & 0.49 & 0.44 \\
    & ROUGE-L & 0.20 & 0.33 & 0.29 & 0.29 & 0.32\\
    \cmidrule{2-7}
    \ldelim\{{1}{*}[trained] & SIDE & 0.40 & 0.36 & 0.37 & 0.26 & 0.10\\
    \cmidrule{2-7}
    \ldelim\{{3}{*}[embedding] & SentenceBERT & 0.36 & 0.47 & 0.52 & 0.56 & 0.41\\
     & \textit{gte-base-en} & 0.39 & 0.52 & 0.55 & 0.57 & 0.46 \\
     & \textit{voyage-code-3} & 0.44 & 0.58 & 0.62 & 0.59 & 0.49 \\
    \cmidrule{2-7}
    \ldelim\{{5}{*}[ask-LLM] & \textit{ask-OLMo} & 0.37 & 0.50 & 0.58 & 0.50 & 0.49 \\
     & \textit{ask-OLMo-no-ref} & 0.38 & 0.55 & 0.59 & 0.45 & 0.50 \\
     & \textit{ask-gpt} & 0.41 & 0.55 & 0.60 & 0.52 & 0.49 \\
     & \textit{ask-claude} & 0.47 & 0.54 & 0.62 & \textbf{0.61} & \textbf{0.62} \\
     & \textit{ask-claude-no-ref} & \textbf{0.48} & \textbf{0.60} & \textbf{0.69} & 0.55 & 0.58\\
    \cmidrule[\heavyrulewidth]{2-7}
    \end{tabular}
    \caption{Spearman's Correlation with Human Ratings for Adequacy and Accuracy}
    \label{table-content-quality}
\end{table*}

\begin{table*}[t]
    \centering
    \small
    \renewcommand{\arraystretch}{0.9}
    \setlength{\tabcolsep}{1pt}
    \begin{tabular}{@{}r@{\,}lccccccc@{}}
    \cmidrule[\heavyrulewidth]{2-9}
     \multicolumn{2}{c}{} & \multicolumn{4}{c}{Conciseness} & \multicolumn{3}{c}{Fluency}\\
     \cmidrule(lr){3-6} \cmidrule(lr){7-9}
     &  && & (Java) & (Python) & & (Java) & (Python)\\
     & Metrics &(\citeauthor{roy2021}) & (\citeauthor{haque2022}) & (\citeauthor{gao2023}) & (\citeauthor{gao2023})& (\citeauthor{roy2021}) &(\citeauthor{gao2023}) & (\citeauthor{gao2023}) \\
    \cmidrule{2-9}
    \ldelim\{{3}{*}[n-gram]
    & BLEU-A & 0.15 & 0.16 & 0.20 & 0.10 & 0.13 & 0.32 & 0.12\\
    & METEOR & 0.17 & 0.27 & 0.19 & 0.23 & 0.15 & 0.29 & 0.23 \\ 
    & ROUGE-L & 0.11 & 0.13 & 0.13 & 0.20 & 0.06 & 0.22 & 0.21\\
    \cmidrule{2-9}
    \ldelim\{{1}{*}[trained] & SIDE & 0.30 & 0.22 & 0.11 & 0.00 & 0.21 & 0.13 & -0.01\\
    \cmidrule{2-9}
    \ldelim\{{3}{*}[embedding] & SentenceBERT & 0.22 & 0.26 & 0.16 & 0.20 & 0.17 & 0.26 & 0.18 \\
     & \textit{gte-base-en} & 0.25 & 0.33 & 0.21 & 0.25 & 0.18 & 0.30 & 0.27\\
     & \textit{voyage-code-3} & 0.29 & 0.37 & 0.16 & 0.25 & 0.22 & 0.27 & 0.26\\
    \cmidrule{2-9}
    \ldelim\{{5}{*}[ask-LLM] & \textit{ask-OLMo} & 0.25 & 0.40 & \textbf{0.33} & 0.32 & 0.17 & 0.37 & 0.32\\
     & \textit{ask-OLMo-no-ref} & 0.28 & 0.41 & 0.31 & 0.33 & 0.22 & 0.31 & 0.33\\
     & \textit{ask-gpt} & 0.29 & 0.42 & 0.32 & 0.28 & 0.24 & 0.34 & 0.31\\
     & \textit{ask-claude} & 0.33 & 0.42 & 0.28 & \textbf{0.35} & \textbf{0.26} & \textbf{0.38} & \textbf{0.38} \\
     & \textit{ask-claude-no-ref} & \textbf{0.36} & \textbf{0.46} & 0.25 & 0.33 & 0.25 & 0.32 & 0.36\\
    \cmidrule[\heavyrulewidth]{2-9}
    \end{tabular}
    \caption{Spearman's Correlation with Human Ratings for Conciseness and Fluency}
    \label{table:style-ratings}
\end{table*}

While we focus on aligning with overall quality, we also look at how well these metrics align with ratings of the summary content and style.
We find that generally, all metrics, including the new baselines we introduce, tend to align better with the content factors of accuracy and adequacy than the style factors of conciseness and fluency. 
In this section, we did not conduct significance tests since this is not our primary objective, but rather intended to just provide supplementary analysis.

\paragraph{Content Ratings} The content (how accurate and how adequate is the information provided), is one of the most important factors in determining overall summary quality according to human raters. In the \citet{haque2022} dataset, Content Adequacy had the highest correlation with Overall Score (Spearman's correlation of 0.83).
In contrast, style factors like conciseness and fluency had correlations of 0.60 and 0.54 respectively.
In \autoref{table-content-quality}, we can see the overall results for these content based factors across three different datasets.

\begin{figure}
    \centering
    \includegraphics[width=\linewidth]{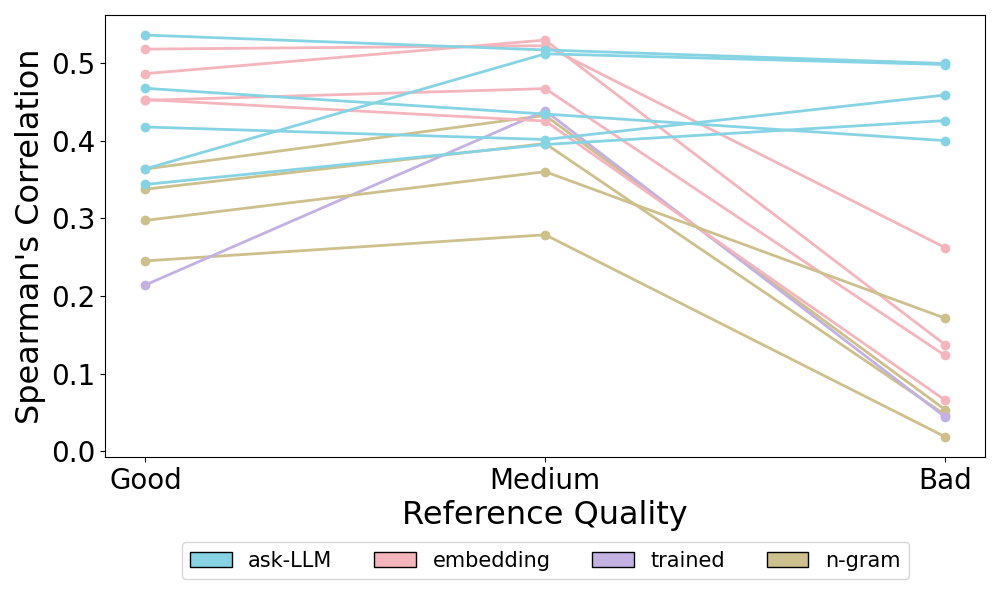}
    \caption{Correlation with Adequacy by Reference Quality on the \citeauthor{haque2022} dataset}
    \label{fig:summary-quality-vs-adequacy}
\end{figure}

Here, the Ask-LLM-Directly approaches with Claude consistently perform better than existing metrics.
One possible cause of this difference is issues with reference summary quality.
\autoref{fig:summary-quality-vs-adequacy} shows how metric performance varies based on the quality of the reference summary in the \citet{haque2022} dataset.
While all of the metrics perform similarly when the references are medium to high quality, performance drops off significantly for all but the ask-LLM metrics when reference quality is low.
One explanation could be that our LLM metrics have access to the code, providing a signal that is independent of the poor quality reference summaries.
Another explanation could be that the n-gram and embedding metrics are focused on similarity, and in this case, high similarity is not an indicator of quality.

\paragraph{Summary Style Ratings}
\autoref{table:style-ratings} shows the correlation with Fluency and Conciseness, which are in general much lower than the correlations with both the overall ratings and the content ratings by around 0.15, which suggests that all of the metrics are prioritising summary content over style when rating. The overall rankings of the metrics remains stable, which means the new improvements in correlation with overall ratings and correlation with adequacy and accuracy are likely not to be coming at the expense of evaluating conciseness and fluency. As metrics continue to improve into the future, it may become useful to develop metrics for the quality factors individually. We attempted to do this by changing the prompt with the Ask-LLM-Directly technique, but we found it was surprisingly difficult to override the LLM's internal representation of overall summary quality by prompting.

\paragraph{Our Metrics Are Effective Across Programming Languages}
\citet{gao2023}'s data includes both Java and Python.
Comparing them in Tables~\ref{table-content-quality} and \ref{table:style-ratings}, we find that our new metrics are fairly stable across languages, while BLEU-A and SIDE are worse on Python.
For SIDE, this is probably because it was trained only on Java.
This flexibility is a strength of the baselines we define.

\begin{table}
    \centering
    \small
    \renewcommand{\arraystretch}{0.9}
    \setlength{\tabcolsep}{1pt}
    \begin{tabular}{@{}r@{\,}lc@{}}
    \cmidrule[\heavyrulewidth]{2-3}
    & & Informativeness\\
     & Metrics & (\citeauthor{su2024contextawarecodesummarygeneration})\\
    \cmidrule{2-3}
    \ldelim\{{3}{*}[n-gram]
    & BLEU-A  & 0.07 \\
    & METEOR & -0.09\\
    & ROUGE-L & 0.22\\
    \cmidrule{2-3}
    \ldelim\{{1}{*}[trained] & SIDE & \textbf{0.45}\\
    \cmidrule{2-3}
    \ldelim\{{3}{*}[embedding] & SentenceBERT  & 0.14\\
     & \textit{gte-base-en} & 0.35 \\
     & \textit{voyage-code-3} & 0.34 \\
    \cmidrule{2-3}
    \ldelim\{{5}{*}[ask-LLM] & \textit{ask-OLMo} & 0.30\\
     & \textit{ask-OLMo-no-ref} & 0.29\\
     & \textit{ask-gpt} & 0.26\\
     & \textit{ask-claude} & 0.23\\
     & \textit{ask-claude-no-ref}  & 0.28\\
    \cmidrule[\heavyrulewidth]{2-3}
    \end{tabular}
    \caption{Spearman's Correlation with Human Ratings for Informativeness}
    \label{informativeness-table}
\end{table}

\paragraph{Correlation with Informativeness}
The results for the final dataset we included, the \citet{su2024contextawarecodesummarygeneration} dataset, are in \autoref{informativeness-table}. It differs from the other datasets we used in a few ways: it includes evaluations of LLM-generated summaries, it evaluates only on the `informativeness' of the summary, and all of the references were individually written and validated as part of another study. Here we see that SIDE actually outperforms all other metrics, but the other commonly used metrics such as the n-gram metrics and the SentenceBERT embeddings-based metric perform particularly poorly, with METEOR even giving a negative correlation. The new embeddings metrics and the LLM-based metrics we introduce both perform similarly, but the correlation is weak-to-moderate overall.
It is clear that this data and/or the human evaluations are measuring something quite different from the other aspects we consider above.
We believe this is because the reference summaries were written with the intent to explain the role of a function in a larger project, rather than explaining what it does.

\subsection{Confidence Intervals}
\label{confidence-intervals-discussion}

We measured confidence intervals and statistical significance by applying the methods from \citet{deutsch-etal-2021-statistical}.
Statistical significance has already been included in discussion above.
For confidence intervals, their approach only applies when there are multiple systems, and so we can only use it for two of the datasets \citep{gao2023,su2024contextawarecodesummarygeneration}.
Due to the small size of these two datasets, the intervals were large.
Note that even with broad confidence intervals, results can be statistically significantly different\footnote{Also note that we only considered significance on the datasets with Overall Score and Similarity, and those datasets are not amenable to this CI calculation method.}.

\section{Significance Testing} 
\label{sec:significance}

\paragraph{Confidence Intervals} \label{confidence-intervals}
We follow the \textsc{Boot-Both}\footnote{Using the implementation from \href{https://github.com/danieldeutsch/nlpstats}{nlpstats}} method from \citet{deutsch-etal-2021-statistical} to calculate confidence intervals, which has been developed specifically for text summarisation metric evaluation. It accounts for human rating data not falling into the normal distribution as well as the fact that we have summaries generated by different models for the same document (code snippet). This approach requires the dataset to have generated summaries from different systems, which meant that it was not possible to calculate confidence intervals for the \citet{haque2022} dataset (they only test one system) and the \citet{roy2021} dataset (which system generated each summary is not included in their publicly available data). 

\paragraph{Permutation Tests}
While \citet{deutsch-etal-2021-statistical} also present code to run p-tests, we did not use their implementation due to the limitation of requiring generated summaries from different systems which we need for two of our datasets.
We implement the test ourselves, sampling 10,000 times to approximate the distribution.
We apply the Bonferroni correction, with 9 p-tests performed on each of the \citet{roy2021} and \citet{haque2022} datasets.

P-values for the tests we ran are shown in \autoref{tab:p-values}.
P-values in bold are significant with p < 0.05 after being adjusted with Bonferroni correction. 

\begin{table*}
    \small
    \centering
    \begin{tabular}{lp{4cm}p{4cm}}
    \toprule
    &  Overall Score & Similarity\\
    & \cite{roy2021} & \cite{haque2022} \\
    \midrule
    \textit{SIDE / best n-gram} & SIDE / BLEU & SIDE / METEOR\\
    & \textbf{0.0002} & \textbf{0.0001}\\
    \midrule
    \textit{SIDE / best embedding} & SIDE / voyage-code-3 & SIDE / voyage-code-3 \\
    & 0.0445 & \textbf{0.0000}\\
    \midrule
    \textit{SIDE / best ask-LLM} & SIDE / ask-claude &  SIDE / ask-claude-no-ref \\
    & 0.0030 & 0.0034\\
    \midrule
    \textit{best embedding / worst embedding} & voyage-code-3 / SentenceBERT & voyage-code-3 / SentenceBERT \\
    & \textbf{0.0000} & \textbf{0.011} \\
    \midrule 
    \textit{best embedding / best n-gram} & voyage-code-3 / BLEU & voyage-code-3 / METEOR \\
    & \textbf{0.0000} & 0.0593\\
    \midrule
    \textit{best embedding / best ask-LLM}& voyage-code-3 / ask-claude & ask-claude-no-ref / voyage-code-3 \\
    & 0.1035 & \textbf{0.0001} \\
    \midrule
    \textit{best ask-LLM / worst ask-LLM} & ask-claude / ask-OLMo & ask-claude-no-ref / ask-OLMo \\
    & \textbf{0.0000} & 0.1344 \\
    \midrule
    \textit{best ask-LLM / best n-gram} & ask-claude / BLEU & ask-claude-no-ref / METEOR \\
    & \textbf{0.0000} & 0.0321 \\
    \midrule
    \textit{ask-claude / ask-claude-no-ref} & ask-claude / ask-claude-no-ref & ask-claude / ask-claude-no-ref \\
    & 0.6779 & 0.3585\\
    \bottomrule
    \end{tabular}
    \caption{P-values for Overall Quality Scores}
    \label{tab:p-values}
\end{table*}

\section{Metrics Tested}
\label{sec:reproduction instructions}

\tightparagraph{BLEU} \cite{papineni-etal-2002-bleu} We use the BLEU-A variant, which is the average for the BLEU score of 1-, 2-, 3- and 4-grams individually.
Calculated using HuggingFace's \texttt{evaluate} package (\url{https://huggingface.co/spaces/evaluate-metric/bleu}). 

\tightparagraph{METEOR} \cite{banerjee-lavie-2005-meteor} is also an n-gram based metric, but gives credit to synonyms and is more highly weighted towards recall.
Calculated using HuggingFace's \texttt{evaluate} package (\url{https://huggingface.co/spaces/evaluate-metric/meteor}).

\tightparagraph{ROUGE-L} \cite{lin-2004-rouge} returns a score based on the longest common subsequence of words in the two summaries. 
Calculated using HuggingFace's \texttt{evaluate} package (\url{https://huggingface.co/spaces/evaluate-metric/rouge}) - the `\texttt{rougeL}' statistic. 

\tightparagraph{SIDE} \cite{mastropaolo2024} uses contrastive learning to train an evaluator model. 
We use the example code provided in \cite{mastropaolo2024}. We used the \texttt{models\_with\_hard\_negatives} version of the model.

\tightparagraph{SentenceBERT} \cite{reimers-gurevych-2019-sentence} is a text embedding method. We apply it to the generated summary and the reference, then calculate cosine similarity.
Computed cosine similarity with \texttt{stsb-roberta-large} available from HuggingFace \texttt{sentence-transformers} (\url{https://huggingface.co/sentence-transformers/stsb-roberta-large}).

\tightparagraph{Claude-3-Opus} We used Claude-3-Opus-20240229, currently available from \url{https://www.anthropic.com/api}.

\tightparagraph{GPT-4o} We used GPT-4o-2024-05-13, currently available from \url{https://platform.openai.com/}.

\tightparagraph{OLMo-2} We used OLMo-2-1124-13B-Instruct, available from HuggingFace (\url{https://huggingface.co/allenai/OLMo-2-1124-13B-Instruct}). 

\tightparagraph{\textit{gte-base-en}} \cite{zhang2024mgtegeneralizedlongcontexttext} is an open-source model which performs well on the Massive Text Embedding Benchmark (MTEB) \cite{muennighoff-etal-2023-mteb} and is small enough to run without a GPU.
on a standard personal computer.
Score calculated with cosine similarity, using \texttt{gte-base-en-v1.5} available on HuggingFace (\url{https://huggingface.co/Alibaba-NLP/gte-base-en-v1.5}). 

\tightparagraph{\textit{voyage-code-3}} \cite{voyagecode3} is a commercial embedding model trained specifically for code.
Computed cosine similarity with VoyageAI's \texttt{voyage-code-3} embedding model as of December 2024. 

\section{Dataset Statistics}
\label{sec:dataset statistics}

\begin{figure}[t]
    \footnotesize
    \textbf{Generated:} \texttt{returns the label text for the given element} \\
    \textbf{Original Code with Reference Summary:} \begin{lstlisting}[language=java, linewidth=\linewidth]
// removes namespace prefix from label text
public String getLabelText(String xpath, String siblingPath, String indexId) {
  if (siblingPath != null && indexId != null) {
    String nodeName = NamespaceRegistry.stripNamespacePrefix(XPathUtils.getNodeName(siblingPath));
    return (nodeName + " ${" + indexId + "+1}");
  } else {
     String nodeName = NamespaceRegistry.stripNamespacePrefix(XPathUtils.getNodeName(normalizedXPath));
     return nodeName;
  }
}
    \end{lstlisting}
    \textbf{Human Ratings:}\\ 
    \begin{tabular}{l l}
        Similarity: & \texttt{[2, 2, 1]} \\
        Accuracy: & \texttt{[3, 4, 4]} \\
        Adequacy: & \texttt{[3, 4, 2]} \\
        Conciseness: & \texttt{[4, 4, 3]} \\
    \end{tabular}
    \caption{Example from \citeauthor{roy2021} (Note: this is a particularly short example)}
    \label{fig:dataset-example}
\end{figure}

Figure~\ref{fig:dataset-example} shows an example of code, a generated summary, and a reference summary.
At the bottom are human annotation of four aspects of quality for the summary.

\subsection{License Information}
The \citeauthor{roy2021} dataset has been released under an MIT license. The Github repositories for the \citeauthor{gao2023} and \citeauthor{haque2022} datasets do not contain any license information. The \citeauthor{su2024contextawarecodesummarygeneration} datasets has not yet been publicly released so they do not have any license yet. Our use is compatible with the intended use when it was provided.

\subsection{Dataset Measures}
\label{sec:data-measures}
\paragraph{\citeauthor{roy2021}} rate from Strongly Agree to Strongly Disagree:

\begin{itemize}
    \setlength{\itemsep}{-3pt}
    \item \textbf{Content Adequacy}: The extent to which the summary lacks information needed to understand the code.
    \item \textbf{Conciseness}: The degree to which the summary contains unnecessary information.
    \item \textbf{Fluency}: The continuity or smoothness rate in the generated summary.
    \item \textbf{Overall Score}: a Direct Assessment (DA) score from 1-100 of the overall quality of the summary.
\end{itemize}

\paragraph{\citeauthor{gao2023}} also rate adequacy, conciseness and fluency, with slightly different definitions from 1 (`very dissatisfied') to 5 (`very satisfied'). These are defined as:

\begin{itemize}
    \setlength{\itemsep}{-3pt}
    \item \textbf{Adequacy}: How much the functional meaning of the code is preserved after summarisation.
    \item \textbf{Conciseness}: The ability to express the function of the code snippet without unnecessary words.
    \item \textbf{Fluency}: The quality of the generated language such as the correctness of grammar.
\end{itemize}

\paragraph{\citeauthor{haque2022}} first show each rater either the reference or generated summary. They rate that summary on accuracy, adequacy and conciseness from Strongly Disagree to Strongly Agree and then rate similarity after seeing the other summary. 

\begin{itemize}
    \setlength{\itemsep}{-3pt}
    \item \textbf{Accuracy}: Independent of other factors, I feel the summary is accurate.
    \item \textbf{Adequacy}: The summary is missing important information, and that can hinder the understanding of the method.
    \item \textbf{Conciseness}: The summary contains a lot of unnecessary information.\footnote{Note that Adequacy and Conciseness are phrased negatively, such that a Strongly Disagree rating is the most positive response. For readability and consistency throughout this paper these have been flipped so high agreement is positive.}
    \item \textbf{Similarity}: These two comments are similar.
\end{itemize}

\paragraph{\citeauthor{su2024contextawarecodesummarygeneration}} focus instead on how the method fits within the entire project, rating the following prompt from Strongly Agree to Strongly Disagree:

\begin{itemize}
    \setlength{\itemsep}{-3pt}
    \item \textbf{Informativeness}: The summary contains information that helps to understand why the method exists in the project. 
\end{itemize}

\subsection{Models Tested}
Each dataset also includes ratings of summaries generated by different models, which have different characteristics based on the method of generation.

\textbf{\citet{roy2021}} sample methods from the Funcom dataset \cite{leclair2019}. They collect ratings of the human-written reference summaries as well as five other summarisation models, listed below in order of human rater preference: 
\begin{itemize}
    \itemsep0em
    \item \textsc{code2seq} \cite{alon2018codeseq} Encoder-Decoder RNN which represents code as compositional paths over its AST.
    \item \textsc{graph2seq} \cite{xu2018graph2seq} Encoder-Decoder RNN which represents code as a graph.
    \item \textsc{ast-attendgru-fc} \cite{haque2020astattengrufc} Encoder-Decoder RNN with three encoders, one for the textual code data, one for the AST, and one for the `file context' - textual code data from other methods in the same file. 
    \item \textsc{ast-attendgru} \cite{leclair2019} Same as \textsc{ast-attendgru-fc} but without the additional file context information.
    \item \textsc{transformer} \cite{vaswani2017} The original Transformer model with no modifications for code summarisation. 
\end{itemize}
Unfortunately, the data available online for this dataset does not include annotations which specify which summary was produced by which model so we are unable to analyse metric performance by model on this particular dataset.

\textbf{\citet{gao2023}} sample methods instead from the TL-CodeSum dataset \cite{hu2018} for the Java data and from the \citet{wan2018} dataset for the Python data. They also select five different code summarisation models, ordered below by human rater preference on `adequacy', but they do not collect ratings of the reference summaries.
\begin{itemize}
    \itemsep0em
    \item \textsc{SG-Trans} \cite{gao2023} Transformer enhanced with structural information of the input, a graph created based on both local structures, e.g. if the tokens belong to the same statement, and global structures, e.g. if there a data flow between the tokens.
    \item \textsc{GREAT} \cite{Hellendoorn2020GlobalRM} Transformer-based model enhanced with graph representations which encode control flow and data flow relations.
    \item \textsc{NeuralCodeSum} \cite{ahmad-etal-2020-transformer} Transformer with small modifications to attention process for the code summarisation task, with no AST or additional code structure information.
    \item \textsc{Transformer} \cite{vaswani2017} The original Transformer model with no modifications for code summarisation. 
    \item \textsc{CodeTransformer} \cite{zugner2021} Transformer which makes both the code and the AST available. It does this in a programming language agnostic way (i.e. it does not require any language-dependent pipelines such as generating a control flow graph).
\end{itemize}

\textbf{\citet{haque2022}}, like \citet{roy2021} also sample data from the Funcom dataset \cite{leclair2019}, but only ask raters to rate the reference summaries and summaries generated by a single baseline, \textsc{attendgru} \cite{leclair2019}, an encoder-decoder RNN which takes only the textual code data as input. 

\textbf{\citet{su2024contextawarecodesummarygeneration}}, instead of drawing from a large dataset of open-source code, uses human-written reference summaries collected by \citet{bansal2024programmervisualattentioncontextaware} where programmers were asked to summarise the purpose of the method in the project. They evaluate the human references, as well as five different methods for generating code summaries:
\begin{itemize}
    \itemsep0em
    \item \textsc{gpt4-base} \cite{su2024contextawarecodesummarygeneration} Summaries obtained by prompting GPT-4. 
    \item \textsc{gpt4-context} \cite{su2024contextawarecodesummarygeneration} Summaries obtained by prompting GPT-4 given summaries of all the functions that call it in the code (these summaries were also automatically generated by the model).
    \item \textsc{gemini-base} \cite{su2024contextawarecodesummarygeneration} Same as \textsc{gpt4-base}, but Gemini is prompted instead.
    \item \textsc{gemini-context} \cite{su2024contextawarecodesummarygeneration} Same as \textsc{gpt4-context}, but Gemini is prompted instead.
    \item \textsc{jam-ft} \cite{su2024contextawarecodesummarygeneration} Fine-tuned version of \textsc{jam} \cite{su2023jam} based on the outputs of \textsc{gemini-context}.
\end{itemize}

\begin{sidewaystable*}
  \begin{center}
    \caption{Human Evaluation Datasets}
    \label{tab:codesumeval}
    \begin{tabular}{|>{\raggedright\arraybackslash}p{2.5cm}|p{1.5cm}|>{\raggedright\arraybackslash}p{4cm}|p{8cm}|>{\raggedright\arraybackslash}p{5cm}|>{\raggedright\arraybackslash}p{2.5cm}|}
      \hline &&&&&\\
      \textbf{Source} &
      \textbf{Language} &
      \textbf{Size} &
      \textbf{Methodology} &
      \textbf{Evaluator Background} &
      \textbf{Link}\\
      &&&&&\\
      \hline
      &&&&&\\
      \citep{haque2022} &
      Java &
      6,300: 210 summary pairs evaluated 30 times each &
      Rate similarity between generated and reference summary and accuracy, content adequacy, conciseness (4-point scale: strongly agree to strongly disagree).  &
      Professional Java developers with an average of 9.3 years experience x 30 &
      \href{https://github.com/similarityMetrics/similarityMetrics}{github} \\
      &&&&&\\
      \hline
      &&&&&\\
      \citep{roy2021} &
      Java &
      6,894: 2,298 summaries evaluated 3 times each &
      Rate conciseness, fluency, adequacy (1-5), rate overall (0-100). &
      \raggedright Professional developers x 48, academics x 61, graduate students x 87, undergraduate students x 17, others x 13 &
      \href{https://github.com/devjeetr/Re-assessing-automatic-evaluation-metrics-for-source-code-summarisation-tasks}{github}\footnote{A more complete version of the dataset including the original methods and summaries is made available by \cite{mastropaolo2024} at \url{https://github.com/antonio-mastropaolo/code-summarisation-metric/tree/main}} \\
      &&&&&\\
      \hline
      &&&&&\\
      \citep{gao2023} &
      Java, Python &
      3,000: 500 Java and 500 Python summaries evaluated 3 times each &
      Rate conciseness, fluency and adequacy (1-5) &
      Professional developers with >4 years experience x 10 &
      \href{https://github.com/shuzhenggao/SG-Trans}{github} \\
      &&&&&\\
      \hline
      &&&&&\\
      \citep{su2024contextawarecodesummarygeneration} &
      Java &
      2400: 240 summary pairs evaluated 10 times each  &
      Rate informativeness: whether `the summary contains information that helps to understand why the method exists in the project'. (4-point scale: strongly agree to strongly disagree).  &
      UK/US residents with a Computer Science degree, and at least 1 year of Java experience x 60 &
      Direct contact with authors\footnote{They have indicated that they will release the dataset publicly soon.}\\
      &&&&&\\
      \hline
    \end{tabular}
  \end{center}
  
\end{sidewaystable*}

\section{Other Variations}
\label{appendix:variations}

\subsection{Prompt Variations}

\begin{table*}
    \centering
    \small
    \setlength{\tabcolsep}{3.5pt}
    \begin{tabular}{
    >{\raggedright\arraybackslash}p{4cm}|
    >{\raggedright\arraybackslash}p{2.2cm}
    >{\raggedright\arraybackslash}p{1.4cm}
    >{\raggedright\arraybackslash}p{2.9cm}
    *{3}{>{\raggedright\arraybackslash}p{1.3cm}}}
    \toprule
         & Quality Dimension & Role Definition & Response Options & Chain of Thought & Reference Summary & Reference Code\\
    \midrule
         consistency-no-ref & Consistent-1 & \xmark & 1-5 & \xmark & \xmark & \cmark\\
         consistency-no-ref-code & Consistent-1 & \xmark & 1-5 & \xmark & \cmark & \xmark\\
         consistency-1-5 & Consistent-1 & \xmark & 1-5 & \xmark & \cmark & \cmark \\
         consistency-0-100 & Consistent-1 & \xmark & 0-100 & \xmark & \cmark & \cmark\\
         consistency-agree-disagree & Consistent-2 & \xmark & Agree/Disagree & \xmark & \cmark & \cmark \\
         accuracy & Accurate-POS & \xmark & Agree/Disagree & \xmark & \cmark & \cmark\\
         adequacy-neg & Adequate-NEG & \xmark & Agree/Disagree & \xmark & \cmark & \cmark \\
         conciseness-neg & Concise-NEG & \xmark & Agree/Disagree & \xmark & \cmark & \cmark \\
         adequacy & Adequate-POS & \xmark & Agree/Disagree & \xmark & \cmark & \cmark \\
         conciseness & Concise-POS & \xmark & Agree/Disagree & \xmark & \cmark & \cmark\\
         accuracy-sftw-eng & Accurate-POS & Soft. Eng. & Agree/Disagree & \xmark & \cmark & \cmark\\
         accuracy-professor & Accurate-POS & Professor & Agree/Disagree & \xmark & \cmark &\cmark \\
         \textbf{Final Method} & Accurate-POS & Soft. Eng. & Agree/Disagree & \cmark & \cmark & \cmark \\
         accuracy-neg & Accurate-NEG & \xmark & Agree/Disagree & \xmark & \cmark & \cmark\\
         accuracy-sftw-eng-cot-neutral & Accurate-POS & Soft. Eng. & Agree/Neutral/Disagree & \cmark & \cmark & \cmark \\
         accuracy-sftw-eng-cot-no-ref & Accurate-POS & Soft. Eng. & Agree/Disagree & \cmark & \xmark & \cmark\\
         informative-sftw-eng-cot & Informative-1 & Soft. Eng. & Agree/Disagree & \cmark & \cmark & \cmark \\
         informative2-sftw-eng-cot & Informative-2 & Soft. Eng. & Agree/Disagree  & \cmark & \cmark & \cmark \\
    \bottomrule
    \end{tabular}
    \caption{Variants tested for Ask-Claude}
\end{table*}

\begin{table*}
    \centering
    \small
    \begin{tabular}{>{\raggedright\arraybackslash}p{6cm}|*{3}{>{\raggedright\arraybackslash}p{2cm}}}
    \toprule
          & Adequacy & Conciseness & Fluency \\
    \midrule
         consistency-no-ref & 0.59 & 0.32 & 0.40\\
         consistency-no-ref-code & 0.55 & 0.31 & 0.36\\
         consistency-1-5 & 0.58 & 0.35 & 0.40 \\
         consistency-0-100 & 0.59 & 0.30 & 0.37\\
         consistency-agree-disagree & 0.58 & 0.41 & 0.42\\ 
         accuracy & 0.59 & 0.38 & 0.43 \\
         adequacy-neg & 0.46 & 0.27 & 0.31\\
         conciseness-neg & -0.37 & -0.32 & -0.40\\
         adequacy-pos & 0.60 & 0.33 & 0.37\\
         conciseness & 0.59 & 0.35 & 0.41\\
         accuracy-sftw-eng & 0.60 & 0.37 & 0.43\\
         accuracy-professor & 0.58 & 0.37 & 0.43\\
         \textbf{Final Method} & 0.64 & 0.32 & 0.43\\
         accuracy-neg & 0.16 & -0.01 & 0.03\\
         accuracy-sftw-eng-cot-neutral & 0.60 & 0.34 & 0.40\\
         accuracy-sftw-eng-cot-no-ref & 0.60 & 0.31 & 0.34\\
    \bottomrule
    \end{tabular}
    \caption{Variants tested for Ask-Claude: Spearman's Correlation with Adequacy, Conciseness and Fluency on \citeauthor{gao2023} training dataset}
\end{table*}

\begin{table}
    \centering
    \small
    \begin{tabular}{lc}
    \toprule
         & Informativeness\\
    \midrule
         \textbf{Final Method} & 0.30\\
         informative-sftw-eng-cot & 0.29\\
         informative2-sftw-eng-cot & 0.23\\
    \bottomrule
    \end{tabular}
    \caption{Quality Dimension Variants tested for Ask-Claude on \citeauthor{su2024contextawarecodesummarygeneration} dataset training dataset}
    \label{table:informativeness-comp}
\end{table}

\begin{figure}
\small
    \begin{mdframed}[skipabove=0pt, skipbelow=0pt, innertopmargin=5pt, innerbottommargin=5pt, innerleftmargin=5pt, innerrightmargin=5pt]
        \texttt{You are a professional software engineer. Evaluate the statement by responding `Strongly agree', `Somewhat agree', `Somewhat disagree' or `Strongly disagree'. Independent of other factors, I feel the new summary is accurate.\\ \\ Reference summary: \{Reference Summary\}\\ Function:\{Original Function\}\\ Generated summary: \{Generated Summary\}\\ What are the steps you would take to evaluate this statement? Show your steps and then provide an evaluation (Strongly agree, Somewhat agree, Somewhat disagree or Strongly disagree).}
    \end{mdframed}
    \caption{Ask LLM Directly Final Prompt}
    \label{ref:llm-prompt}
\end{figure}

Figure~\ref{ref:llm-prompt} shows the final prompt used.
In the reference free case, the "Reference Summary" line is left out.
    
We varied the Ask Claude Directly prompts in six different ways: the quality dimension definition the summary was to be rated on, the role definition for role-based prompting, the format of the expected response, whether chain-of-thought prompting was used, whether the reference summary was included and whether the reference code was included. 

\subsubsection{Quality Dimensions}
\paragraph{Consistent-1} Rate how consistent the following summary is with the corresponding function and reference summary. Note that consistency means that all the information in the new summary is supported by the code [or the reference summary, when provided].
\paragraph{Consistent-2} The following summary is consistent. Note that consistency means that all the information in the new summary is supported by the code [or the reference summary, when provided].
\paragraph{Accurate-POS} Independent of other factors, I feel the new summary is accurate.
\paragraph{Accurate-NEG} Independent of other factors, I feel the new summary is inaccurate.
\paragraph{Adequate-POS} The new summary contains all of the important information required for understanding the method.
\paragraph{Adequate-NEG} The new summary is missing important information, and that can hinder the understanding of the method.
\paragraph{Concise-POS} The new summary only contains necessary information.
\paragraph{Concise-NEG} The new summary contains a lot of unnecessary information.
\paragraph{Informative-1} The summary contains information that helps to understand why the method exists in the project
\paragraph{Informative-2} Independent of other factors, I feel that the new summary contains relevant information that helps to understand why the method exists in the project

\subsubsection{Role Definitions}
\paragraph{Software Engineer} You are a professional software engineer.
\paragraph{Professor} You are a Professor of Computer Science at a reputable university.

\subsubsection{Response Options}
We often mentioned the response options multiple times in the prompt. In italics is the location of the that particular piece of text which can be cross-referenced with the prompt scaffolds in \autoref{sec:prompt-scaffolds}.
\paragraph{1-5} \textit{After Data} Rating (1 to 5):
\paragraph{0-100} \textit{Before Data} Give a rating from 0 to 100 where 0 means completely inconsistent and 100 means the summary is fully consistent with the code or the reference summary. \\
\textit{After Data} Rating (0 to 100):
\paragraph{Agree-Disagree} \textit{Before Quality Dimension} Evaluate the statement by responding `Strongly agree', `Somewhat agree', `Somewhat disagree' or `Strongly disagree'.\\
\textit{After Data} Evaluation (Strongly agree, Somewhat agree, Somewhat disagree or Strongly disagree):
\paragraph{Agree-Disagree + Chain of Thought} \textit{Before Quality Dimension} Evaluate the statement by responding `Strongly agree', `Somewhat agree', `Somewhat disagree' or `Strongly disagree'.\\
\textit{After Data} What are the steps you would take to evaluate this statement? Show your steps and then provide an evaluation (Strongly agree, Somewhat agree, Somewhat disagree or Strongly disagree):
\paragraph{Agree-Neutral-Disagree + Chain of Thought}  \textit{Before Quality Dimension} Evaluate the statement by responding (Strongly agree, Somewhat agree, Neutral, Somewhat disagree or Strongly disagree).\\
\textit{After Data} What are the steps you would take to evaluate this statement? Show your steps and then provide an evaluation (Strongly agree, Somewhat agree, Neutral, Somewhat disagree or Strongly disagree):

\subsubsection{Data Provided}
\paragraph{Reference Summary and Reference Code} \mbox{}\\
Reference summary:\{reference summary\}\\ Function:\\\{reference method\}\\
Generated summary: \{generated summary\}
\paragraph{Reference Summary Only} \mbox{}\\
Reference summary: \{reference summary\} \\
Generated summary: \{generated summary\}
\paragraph{Reference Code Only} \mbox{}\\
Function:\\\{reference method\}\\
Generated summary: \{generated summary\}

\subsubsection{Prompt Scaffold}
\label{sec:prompt-scaffolds}
\begin{flushleft}
\texttt{[Role Definition] [Response Options: Before Quality Dimension] [Quality Dimension] [Response Options: Before Data]}\\
\texttt{[Data Provided]} \\
\texttt{[Response Options: After Data]}
\end{flushleft}

\subsection{Question-Answering Variation}
We also tried to use Claude as part of a Question-Answering-style metric, inspired by previous work in text summarisation (e.g. QAGS \cite{wang-etal-2020-asking}, QAEval \cite{deutsch-etal-2021-towards}, QuestEval \cite{scialom-etal-2021-questeval} and QAFactEval \cite{fabbri-etal-2022-qafacteval}), but this approach did not end up providing any improvements compared to the n-gram based metrics. The main idea is that after reading a good generated summary you should be able to answer questions about the subject similarly to if you had read the reference summary instead. The details of our approach are as follows: 

\begin{enumerate}
\setlength{\itemsep}{-3pt}
\item Find all noun phrases in the reference summary using spaCy (en\_core\_web\_sm).
\item Generate questions by replacing each noun phrase with a gap. For example, for the summary \texttt{``returns the label text for the given element''}, one question would be \texttt{``returns the \_ \_ \_ for the given element''}. 
\item Given only the generated summary, prompt an LLM to try to fill in the blank for each question generated in Step 2. We used Claude 3 Opus.
\item Compare the correct answers with the responses generated by the model by converting each answer to an embedding (we used gte-base-en), and calculating cosine similarity.
\item Return the mean of the cosine similarity scores for each question as the final score. 
\end{enumerate}

The main difference between our approach and the standard approach for text summarisation is Step 2, question generation, as we programmatically generate the questions as a fill-in-the-blank rather than ask an LLM to generate the questions given the reference code and summary. The reason for this change was because we found that the questions generated were too specific (e.g., `What does the variable \texttt{i} do?'), whose answers don't appear in a code summary.

\begin{table*}[ht]
    \centering
    \small
    \setlength{\tabcolsep}{3.5pt}
    \begin{tabular}{>{\raggedright\arraybackslash}p{3cm}|*{6}{>{\raggedright\arraybackslash}p{1.5cm}}>{\raggedright\arraybackslash}p{2cm}}
    \toprule
          & +verb phrases & +few shot & +return as word only & +n.a. if not enough information & +use whole result sentence for similarity & +use gte-base-en for embeddings & +handle n.a. differently  \\
    \midrule
         \textbf{Final QA Method} & \xmark & \cmark & \cmark & \xmark & \xmark & \cmark &n.a.\\
         NA-counts-as-0 & \xmark &\cmark & \cmark & \cmark & \xmark & \cmark & \cmark (n.a. = 0)\\
         NPs-Only & \xmark & \cmark & \cmark & \cmark & \xmark & \cmark & \xmark \\
         NPs-Only-JSON & \xmark & \cmark & \xmark & \cmark & \xmark & \cmark & \xmark\\
         NA-counts-as-0.5 & \xmark & \cmark & \cmark & \cmark & \xmark & \cmark & \cmark (n.a. = 0.5) \\ 
         NA-not-counted &\xmark &\cmark & \cmark & \cmark & \xmark &\cmark &\cmark (exclude n.a.) \\
         NPs+VPs & \cmark & \cmark & \cmark & \cmark & \xmark & \cmark &\xmark\\
         NPs+VPs-SBERT & \cmark & \cmark & \cmark & \cmark &\xmark & \xmark & \xmark \\
         NPs-Only-zero-shot & \xmark & \xmark &\cmark & \cmark & \xmark &\cmark & \xmark \\
         full-sent-SBERT & \cmark & \cmark & \cmark & \cmark & \cmark & \xmark & \xmark\\
         full-sent-GTE & \cmark & \cmark & \cmark & \cmark & \cmark & \cmark & \xmark\\
    \bottomrule
    \end{tabular}
    \caption{Variants tested for Question Answering (Ordered from best to worst on Accuracy on \citeauthor{gao2023} (Java) train dataset)}
    \label{table:variant-defs}
\end{table*}

\begin{table*}
    \centering
    \small
    \begin{tabular}{>{\raggedright\arraybackslash}p{5cm}|*{3}{>{\raggedright\arraybackslash}p{2cm}}}
    \toprule
          & Adequacy & Conciseness & Fluency \\
    \midrule
         \textbf{Final QA Method }& \textbf{0.55} & 0.11 & 0.20 \\
         NA-counts-as-0 & 0.54 & 0.17 & 0.25\\
         NPs-Only & 0.54 & 0.16 & 0.22 \\
         NPs-Only-JSON & 0.53 & 0.17 & 0.25\\
         NA-counts-as-0.5 & 0.53 & 0.14 & 0.20 \\ 
         NA-not-counted & 0.53 & 0.14 & 0.20 \\
         NPs+VPs & 0.53 & \textbf{0.21} & 0.28\\
         NPs+VPs-SBERT & 0.49 & 0.20 & \textbf{0.29}\\
         NPs-Only-zero-shot & 0.46 & 0.13 & 0.22\\
         full-sent-SBERT & 0.45 & 0.11 & 0.22\\
         full-sent-GTE & 0.44 & 0.07 & 0.16\\
    \bottomrule
    \end{tabular}
    \caption{Variants tested for Question Answering: Spearman's Correlation with Adequacy, Conciseness and Fluency on \citeauthor{gao2023} training dataset}
    \label{table:qa-variant-results}
\end{table*}

\subsubsection{QA Prompt}
\begin{figure}[H]
    \small
    \begin{mdframed}[skipabove=0pt, skipbelow=0pt, innertopmargin=5pt, innerbottommargin=5pt, innerleftmargin=5pt, innerrightmargin=5pt]
        \texttt{Based on the following code summary, fill in the blanks for the other code summary based on the same function.\\ \\
For example:\\
Code Summary: `get the list of the user'\\
Question: `returns \_\_\_ of collaborate collections for the given user id'\\
Answer: `the list'\\\\
Code Summary: \{Generated\_summary\}\\
Question: \{Question\}\\
Answer:}
    \end{mdframed}
    \caption{Question Answering Prompt for Question Generation Step}
\end{figure}

\subsubsection{Variants}
We tested many different variants on the \citeauthor{gao2023} dataset, varying seven different aspects of the process. The combinations tested are provided in \autoref{table:variant-defs}, and the results are in \autoref{table:qa-variant-results}. We outline each of the variations below:
\paragraph{+ verb phrases} In the answer selection step, include all tokens whose part of speech is `VERB' as well as all of the `noun chunks', as outputted by \texttt{SPaCY en\_core\_web\_sm}.
\paragraph{- verb phrases} In the answer selection step, only include the `noun chunks' from the summary.
\paragraph{+ few shot} Include an example(s) of the expected output in the prompt. The exact wording depends on 1) the return format and 2) if n.a. is an option if there is not enough information.\\\\
\textit{JSON prompt (with n.a.):}
\begin{verbatim}
Based on the following code summary, 
fill in the blanks for the other 
code summary based on the same function.

For example:
Code Summary: 'get the list of the user'
Question: returns ___ of collaborate 
collections for the given user id
Answer:{ 
    "answers": [
        "the list"
    ]
}

  
\end{verbatim}
\textit{Return as word only prompt (without n.a.):}
\begin{verbatim}
Based on the following code summary, 
fill in the blanks for the other 
code summary based on the same function.

For example:
Code Summary: `get the list of the user`
Question: `returns ___ of collaborate 
collections for the given user id`
Answer: `the list`
\end{verbatim}
\paragraph{- few shot} In the no few shot scenario, the prompt just includes an explanation of the task, e.g. 
\begin{verbatim}
Use only the information from the code 
summary to fill in the blank on the 
following question. If there is not enough 
information to give an answer, write 'n.a.'.
\end{verbatim}
\paragraph{+ return as word only} Prompt the LLM to just return the answer to the question. 
\begin{verbatim}
Only provide your answer in the response. 
... {the summary and the question} ...
Answer:
\end{verbatim}
\paragraph{- return as word only} Prompt the LLM to return the answer to the question in JSON format.
\begin{verbatim}
Return your answer in json format, 
for example
{
    "answers": [
         "answer"
    ]
}
\end{verbatim}
\paragraph{+ n.a. if not enough information} Prompt the LLM to return n.a. if it thinks there is not enough information in the summary to answer the question.  
\paragraph{- n.a. if not enough information} Do not specify how to handle situations where there is not enough information in the summary to answer the question.
\paragraph{+ use whole result sentence for similarity} The response from the LLM was re-inserted back into the blank, and then the embedding of this full sentence was compared against the embedding of the original summary.
\paragraph{- use whole result sentence for similarity} Only the single word that was generated and the original word that should have filled the blank were compared.
\paragraph{+ use gte-base-en for embeddings} Embeddings for each of the generated and reference answers were calculated using \texttt{gte-base-en-v1.5}.
\paragraph{- use gte-base-en for embeddings} Embeddings were instead calculated with \texttt{SentenceBERT}.
\paragraph{+ handle n.a. differently} When the model responded with n.a., these questions were automatically assigned a predetermined similarity score (0 or 0.5) or excluded from the final calculation completely.
\paragraph{- handle n.a. differently} For cases where the model responded with n.a., the contribution of n.a. to the final score was just the cosine similarity of the expected answer and the string `n.a.', i.e. handled the same as all other answers.

\section{Claude Evaluating Itself}
\label{sec:claude-evaluates-itself}
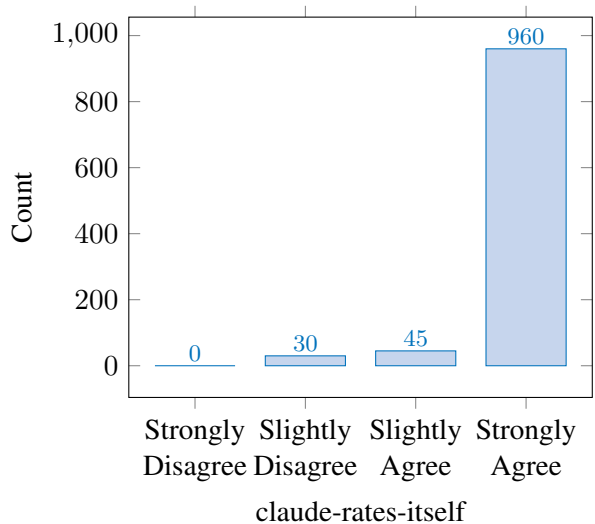
\begin{figure}[H]
\centering
\begin{tikzpicture}
\pgfplotsset{compat=1.18}
\begin{axis}[
        width= 0.48\textwidth,
	ylabel=Count,
	ybar,
        bar width = 30pt,
	xtick=data,
        enlarge x limits=0.2,
	xticklabels={Strongly\\Disagree, Slightly\\Disagree, Slightly\\Agree, Strongly\\Agree},
        xticklabel style = {align=center},
        nodes near coords,
        nodes near coords style={font=\footnotesize, yshift=-2pt},
	xlabel={claude-rates-itself},
]
\addplot[color=RoyalBlue, fill=RoyalBlue!20] 
	coordinates {(1,0) (2,30) (3, 45) (4,960)};
\end{axis}
\end{tikzpicture}
\caption{Ask-Claude scores for summaries generated by Claude 3.5 (for functions from the \citeauthor{gao2023} dataset)}
\label{fig:claude-rates-itself}
\end{figure}

We asked Claude to rate its own output.  The summaries were generated with the prompt in \autoref{claude-generates-summary-prompt}. They were evaluated using the Ask-Claude consistency-agree-disagree prompt on the \citeauthor{gao2023} Java training dataset. The results are in \autoref{fig:claude-rates-itself}.
\begin{figure}[H]
    \small
    \begin{mdframed}[skipabove=0pt, skipbelow=0pt, innertopmargin=5pt, innerbottommargin=5pt, innerleftmargin=5pt, innerrightmargin=5pt]
        \texttt{Write a comment that summarises the following code. Ensure that it is fully consistent, so all information in the comment is supported by the code.\\
        Function: \{Function\}\\
        Comment:}
    \end{mdframed}
    \caption{Prompt given to Claude for Summary Generation}
    \label{claude-generates-summary-prompt}
\end{figure}

\section{Correlation with Comment Length}
\label{sec:corr-comm-leng}

See \autoref{table:comment-length}.
\begin{table*}[ht]
    \centering
    \small
    \renewcommand{\arraystretch}{0.8}
    \begin{tabular}{@{} l *{5}{>{\raggedright\arraybackslash}p{1.75cm}} @{}}
    \toprule
    & \citeauthor{roy2021} & \citeauthor{haque2022} & \citeauthor{gao2023} (Java) & \citeauthor{gao2023} (Python) & \citeauthor{su2024contextawarecodesummarygeneration} \\
    \midrule
    adequacy &  0.00 & 0.15 & -0.04 & 0.00 &  n.a. \\
    conciseness &  -0.08 & -0.08 & -0.36 & -0.02 & n.a. \\
    fluency  & -0.02 & n.a. & -0.15 & 0.06 & n.a. \\
    accurate  & n.a. & 0.06 & n.a. & n.a. & n.a. \\
    similarity & n.a. & 0.27 & n.a. & n.a. &  n.a. \\
    overall score & -0.01 & n.a. & n.a. & n.a. & n.a. \\
    informativeness & n.a. & n.a. & n.a. & n.a.& -0.41 \\
    \midrule
    BLEU-A &  0.20 & 0.43 & -0.02 & 0.05 & -0.07 \\
    METEOR & 0.16 & 0.32 & -0.08 & 0.02  & 0.06 \\
    ROUGE-L &  0.03 & 0.25 & -0.20 & -0.05 & -0.35 \\
    \midrule
    SIDE &  0.02 & 0.16 & -0.21 & -0.01  & -0.49 \\
    \midrule
    SentenceBERT & 0.09 & 0.35  & -0.13 & 0.02 & 0.17 \\
    gte-base-en &  0.07 & 0.29 & -0.12 & 0.01 & -0.03 \\
    voyage-code-3 & 0.06 & 0.33 & -0.10 & 0.00 & -0.01 \\
    \midrule
    ask-OLMo & -0.08 & 0.14& -0.15 & -0.05 & 0.25 \\
    ask-OLMo-no-ref & -0.13 & 0.14 & -0.18 & -0.01 & 0.20 \\
    ask-gpt & -0.03 & 0.15 & -0.24 &  0.06 &  0.22 \\
    ask-claude & -0.01 & 0.13 & -0.12 & -0.01 & 0.26 \\
    ask-claude-no-ref&  -0.11 & 0.13 & -0.19 & -0.03 & 0.25 \\
    \bottomrule
    \end{tabular}
    \caption{Correlation with Comment Length}
    \label{table:comment-length}
\end{table*}

\section{Relative Rankings of Each Model by Metric}
\label{sec:rel-ranks}

\begin{figure*}[ht]
        \subfloat[Model Rankings by Metric (\citeauthor{gao2023} Java, Adequacy)]{%
            \includegraphics[width=.48\linewidth]{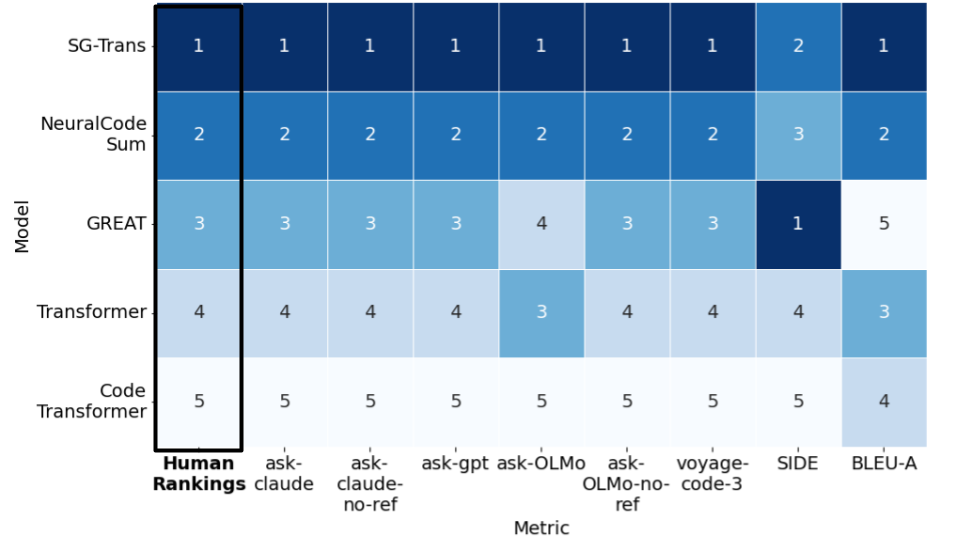}%
            \label{subfig:a}%
        }\hfill
        \subfloat[Model Rankings by Metric (\citeauthor{gao2023} Python, Adequacy)]{%
            \includegraphics[width=.48\linewidth]{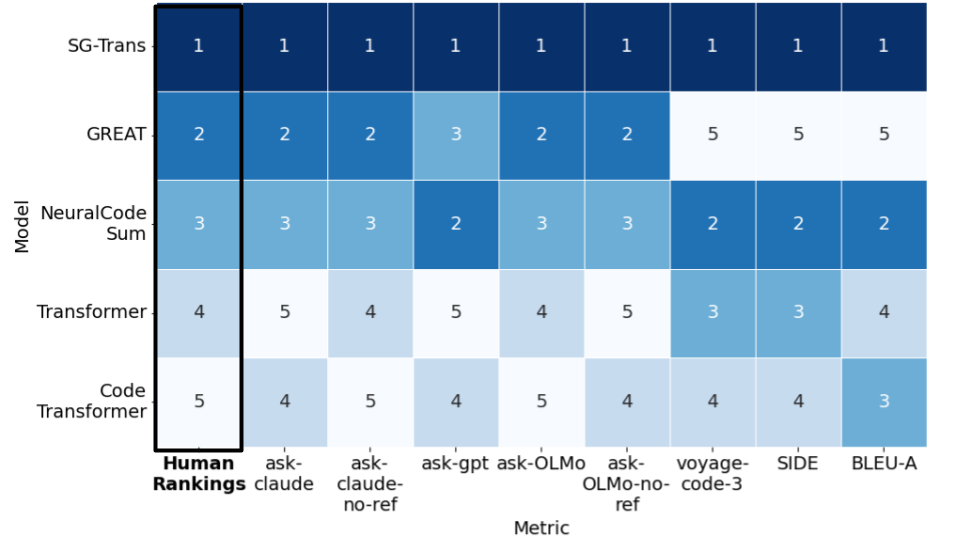}%
            \label{subfig:b}%
        }\\
        \subfloat[Model Rankings by Metric (\citeauthor{su2024contextawarecodesummarygeneration}, Informativeness)]{%
            \includegraphics[width=.48\linewidth]{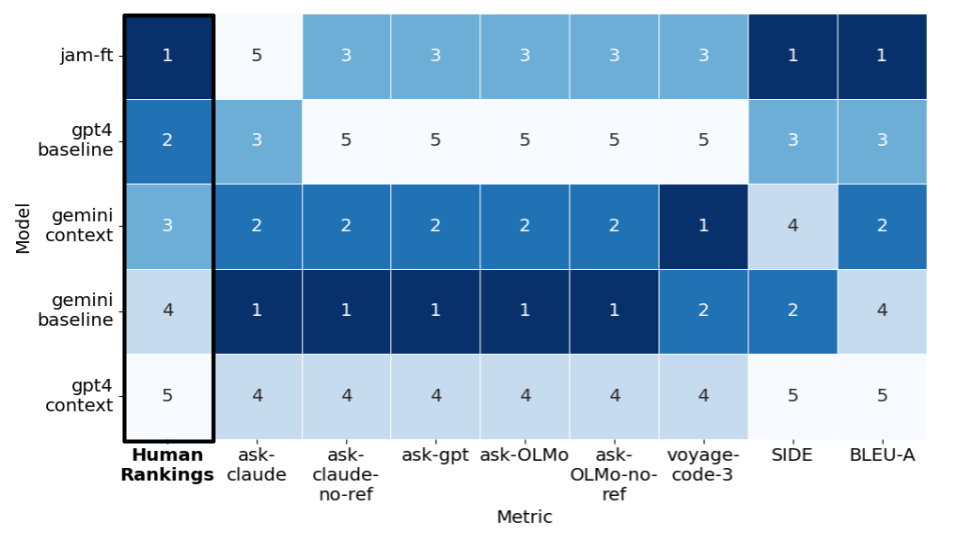}%
            \label{subfig:d}%
        }
        \caption{Relative Rankings of Each Model by Metric}
        \label{fig:relative-rankings}
    \end{figure*}

See \autoref{fig:relative-rankings}.

\end{document}